\documentclass[letterpaper, 10 pt, conference]{ieeeconf}

\IEEEoverridecommandlockouts 

\makeatletter
\let\NAT@parse\undefined
\makeatother

\usepackage[hidelinks, breaklinks]{hyperref}

\hypersetup{
     colorlinks   = false,
     citecolor    = gray,
     linkcolor    = black,
     urlcolor     = gray,
} 

\usepackage{graphics} 
\usepackage{epsfig} 
\usepackage{mathptmx}
\usepackage{times} 
\usepackage{colortbl}
\usepackage{amsmath}
\usepackage{amssymb}
\usepackage{amsfonts}
\usepackage{mathtools}
\usepackage{graphicx}
\usepackage{textcomp}
\usepackage{xcolor}
\usepackage[nolist]{acronym}
\usepackage{acronyms}
\usepackage[nocompress]{cite}
 \usepackage{bm}
\usepackage{dirtree}
\usepackage{balance}
\usepackage[mathscr]{euscript}
\usepackage{wrapfig}
\usepackage{lipsum}
\usepackage{caption}
\usepackage{subcaption}
\usepackage[misc]{ifsym}
\usepackage[T1]{fontenc}
\usepackage{float}
\usepackage{stfloats} 
\usepackage{tabularray}
\usepackage{diagbox}
\usepackage[utf8]{inputenc}
\usepackage{color}
\usepackage{array}     
\usepackage{adjustbox}    
\usepackage{multirow}
\usepackage{orcidlink}   
\usepackage{booktabs}
\usepackage{siunitx}
\usepackage[absolute,overlay]{textpos}
\usepackage{todonotes}

\usepackage{calc} 

\makeatletter
\let\MYcaption\@makecaption
\makeatother
\usepackage[font=footnotesize]{subcaption}
\makeatletter
\let\@makecaption\MYcaption
\makeatother

\newlength{\colA}\newlength{\colB}\newlength{\colC}
\newlength{\rowA}\newlength{\rowB}
\newlength{\rightHeight}
\newlength{\hgap}\newlength{\vgap}
\title{\LARGE \bf Omnidirectional Solid-State mmWave Radar Perception \\ for UAV Power Line Collision Avoidance}

\author{Nicolaj Haarhøj Malle\orcidlink{0000-0002-3158-0509} and \hspace{-4mm}
\thanks{This work was supported by Innovation Fund Denmark's Innoexplorer programme under grant agreement AIR-Ops, the Villum Foundation's Spin-outs Denmark programme under grant agreement OnGrid, and the Carlsberg Foundation under grant agreement AIR-REPAIR.}
\thanks{\noindent All authors are affiliated with the Drone Infrastructure Inspection and Interaction Group, Section for Digital and High-Frequency Electronics, Department of Mechanical and Electrical Engineering, Faculty of Engineering, University of Southern Denmark, \texttt{email: nhma@sdu.dk}}
Emad Ebeid\orcidlink{0000-0002-7847-149X}
}

\begin{document}

\maketitle


\begin{abstract}

Detecting and estimating distances to power lines is a challenge for both human UAV pilots and autonomous systems, which increases the risk of unintended collisions. We present a mmWave radar–based perception system that provides spherical sensing coverage around a small UAV for robust power line detection and avoidance. The system integrates multiple compact solid-state mmWave radar modules to synthesize an omnidirectional field of view while remaining lightweight. We characterize the sensing behavior of this omnidirectional radar arrangement in power line environments and develop a robust detection-and-avoidance algorithm tailored to that behavior. Field experiments on real power lines demonstrate reliable detection at ranges up to 10 m, successful avoidance maneuvers at flight speeds upwards of 10 m/s, and detection of wires as thin as 1.2 mm in diameter. These results indicate the approach’s suitability as an additional safety layer for both autonomous and manual UAV flight.

\end{abstract}

\vspace{0.25cm}
A video demonstration of the system can be viewed in\cite{video_demo}, and instructions and code can be found in\cite{srd_ros2_pkg}.

\section{Introduction}
\label{sec:introduction}
With the proliferation of UAVs in recent years and the continued build-out of power transmission networks around the world, the risk of collisions between UAVs and power lines is increasing. The severity of such a collisions range from mostly harmless, where both UAV and infrastructure remain undamaged, to worse collisions leading to UAV crashes, damage to infrastructure, and potential disruptions to power distribution. 
UAV pilots are challenged by power lines whose distances are often difficult to assess from the ground or through compressed video streams on remote control stations. For autonomous systems, power lines may be hard to detect because of their thin and feature-poor appearance. UAV platforms may also lack omnidirectional sensing, leading to blind spots and increased risk of collisions from certain angles. This is critical for multirotor-type UAVs which can move in any direction and are therefore at greater risk of collisions.

\begin{figure}[t]
    \centering
    \includegraphics[width=0.92\linewidth]{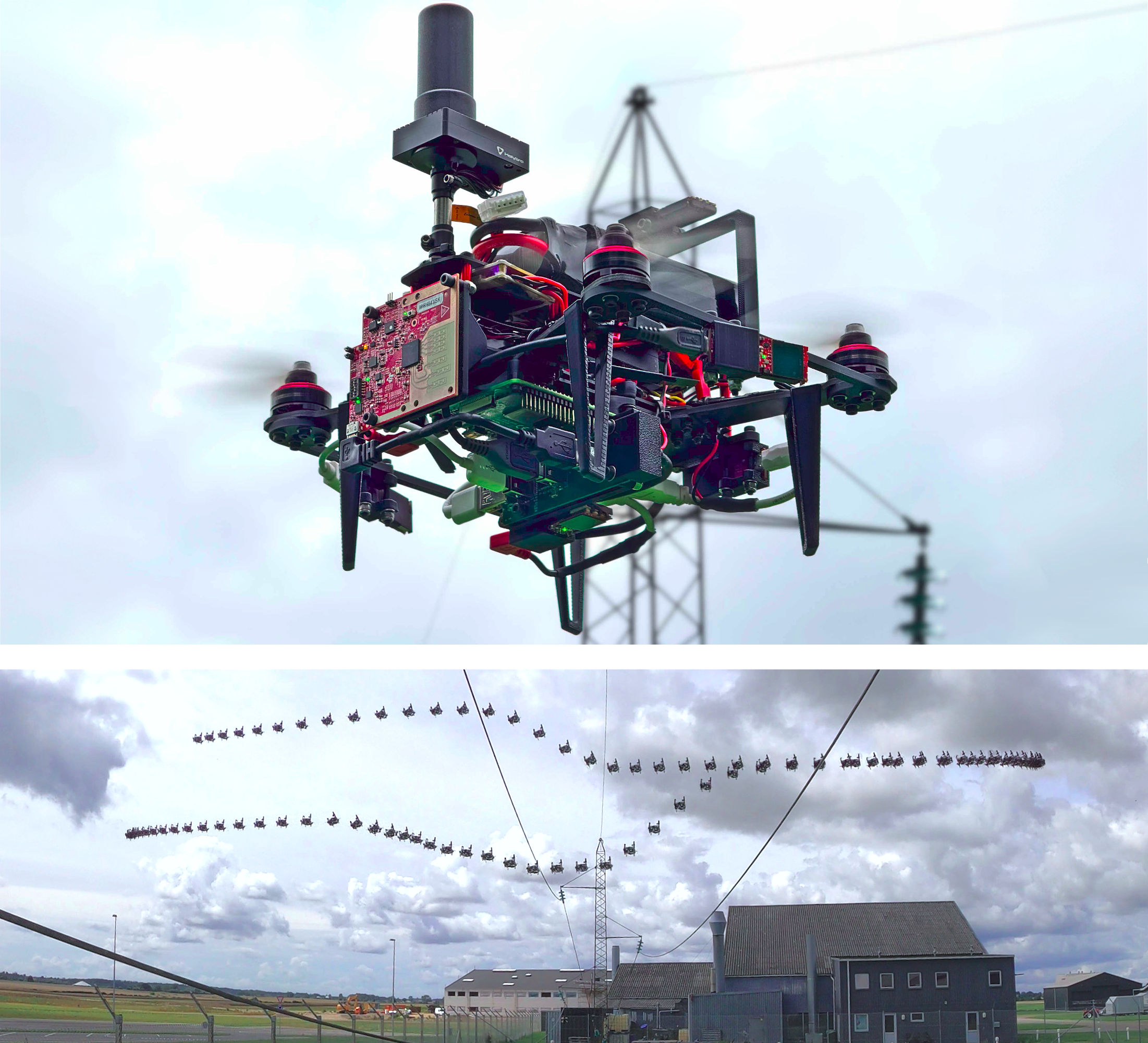}
    \caption{\textbf{Top}: Omnidirectional power line avoidance system flying near power lines with all six radar devices visible. \textbf{Bottom}: Time-lapse of power line avoidance maneuvers.}
    \label{fig:title}
\end{figure}

In this work we propose a multi-sensor millimeter-wave (mmWave) radar-based power line avoidance system with omnidirectional sensing. The system, shown in Fig. \ref{fig:title}, reliably detects power lines from multiple meters away and avoids them while flying at high speeds. The power line avoidance scheme can then be added as a layer of protection on top of either manual or autonomous flight modes.

The contributions of this work are:
\begin{itemize}
    \item A lightweight solid-state multi-radar perception architecture that synthesizes omnidirectional coverage.
    \item A characterization of the sensing behavior of the omnidirectional radar architecture with power line targets.
    \item A robust power line avoidance algorithm that exploits radar's specific interaction with power lines.
\end{itemize}

The structure of the paper is as follows: Sec. \ref{sec:related_work} explores related work; Sec. \ref{sec:scope} outlines the scope of the work; Sec. \ref{sec:hardware} describes the physical system; Sec. \ref{sec:radar_pl_behavior} examines the behavior of radar with power line targets; Sec. \ref{sec:avoidance} describes the avoidance algorithm; Sec. \ref{sec:experiments} shows the flight test results; And Sec. \ref{sec:conclusion} concludes on the work.

\section{Related Work}
\label{sec:related_work}
In previous work we have demonstrated how mmWave radar devices can perform reliable power line detection~\cite{Malle2021} to enable lightweight tracking~\cite{Malle2022} and thereby facilitate physical interaction with and recharging from power lines~\cite{icra2023}~\cite{icra2024} as well as mapping and reconstruction~\cite{iros2024} of power lines using UAVs. All these previous systems have relied on just a single radar sensor to achieve their respective capabilities, thus leaving them effectively blind outside of the sensor field of view. Furthermore, none of these systems implement any kind of obstacle avoidance. Lastly, no characterization of mmWave radar behavior in power line environments has been conducted. This is especially evident in~\cite{Malle2022}, which relies on an additional camera sensor and image processing pipeline to determine the direction of the power lines - information that may instead be extracted directly from the radar data, as shown in this work. 

Other existing UAV platforms also rely on both a camera and radar sensor for obstacle avoidance, such as Yu et al.\cite{9341432} and Bigazzi et al.\cite{drones6110361}, and object detection as in the system by Wang et al.\cite{radarcameracnn}. A few other UAV systems rely solely on radar sensors for tasks such as target detection by Milias et al.\cite{10075053}, 3D reconstruction by Sun et al.\cite{9973561} and generic obstacle avoidance by Wessendorp et al.\cite{wessendorp2021} However, none of these systems have omnidirectional radar sensing and all of them are therefore susceptible to collisions with objects in their blind spots. 
Wu et al.\cite{10772391} attempt to address the lack of omnidirectional sensing by developing a 360° radar device meant for UAV obstacle avoidance. 
The authors mainly evaluate the antenna and beam-switching of their device, but no object detection experiments are carried out. Furthermore, their device only targets the horizontal plane, neglecting anything above and below this plane.
\section{Scope}
\label{sec:scope}

A review of the existing literature reveals a number of shortcomings:
\begin{itemize}
    \item There is an absence of systems that rely purely on radar sensors for long range, all-weather power line detection and avoidance.
    \item Existing work lacks spherical sensing to enable detection and avoidance in any direction.
    \item There has been little work towards analyzing the behavior of mmWave radar sensors in power line environments and subsequently taking advantage of the unique interaction.
    \item There is a lack of testing of existing work in real world power line environments to validate their functionality.
\end{itemize}

To address the shortcomings in existing literature, the scope of the presented work is to achieve spherical radar-only sensing coverage by surrounding the UAV frame with multiple mmWave radar devices. With a functioning platform, the aim is then to characterize how the omnidirectional mmWave radar perception behaves in power line environments and to take advantage of the unique interaction to develop an efficient and lightweight power line avoidance algorithm. The developed system is then extensively tested in both laboratory and outdoors power line environments to evaluate the detect-and-avoid functionality and determine its limits.

\section{Hardware Setup}
\label{sec:hardware}
\begin{figure}[t]
    \centering
    \begin{subfigure}[b]{\linewidth}
        \centering
        \includegraphics[width=0.99\linewidth]{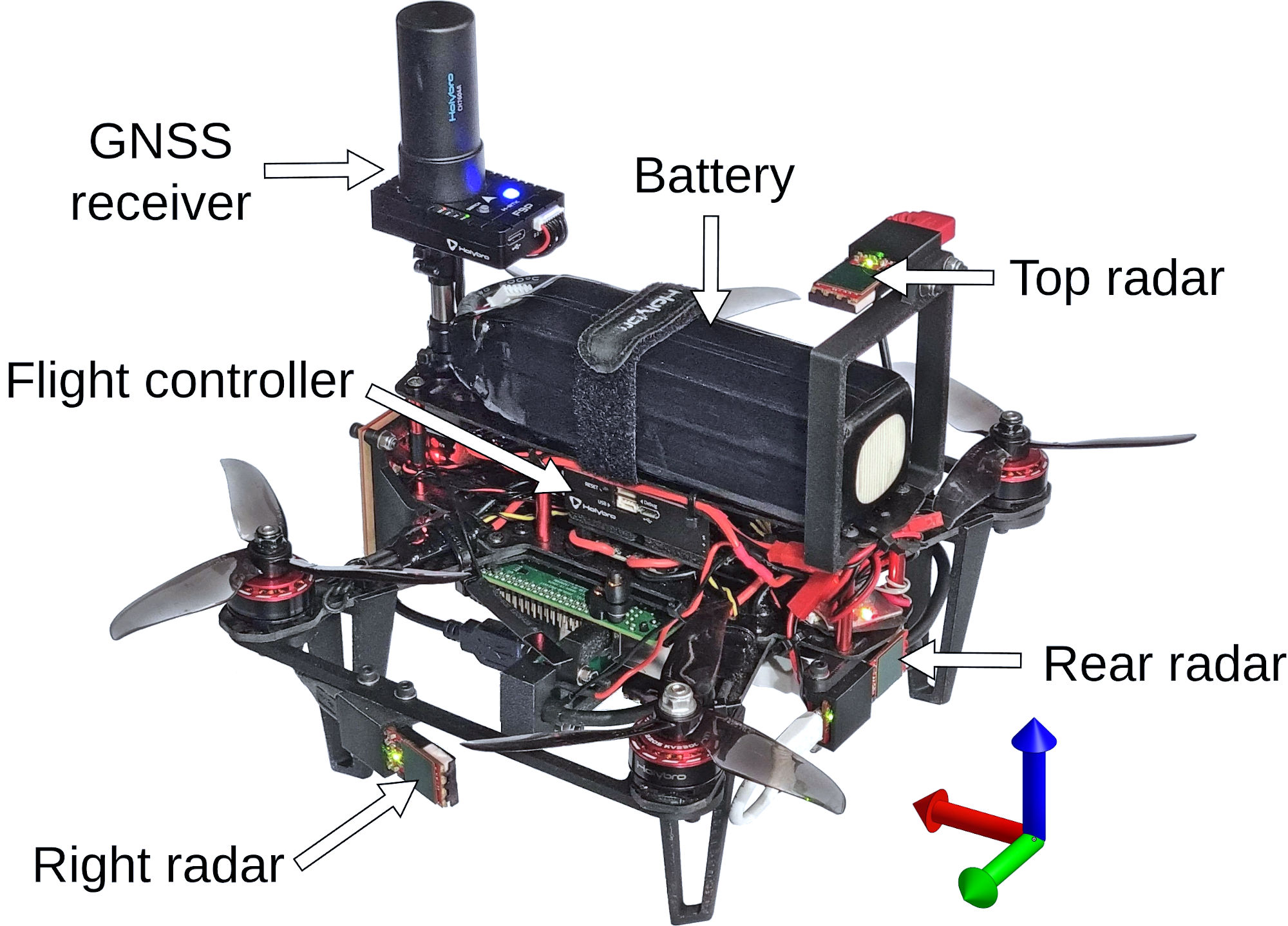}
        \caption{System seen from the the top rear, showing the location of the GNSS receiver, battery, autopilot, and some of the radar sensors.}
        \label{subfig:top}
    \end{subfigure}

    \vspace{1em} 

    \begin{subfigure}[b]{\linewidth}
        \centering
        \includegraphics[width=0.9\linewidth]{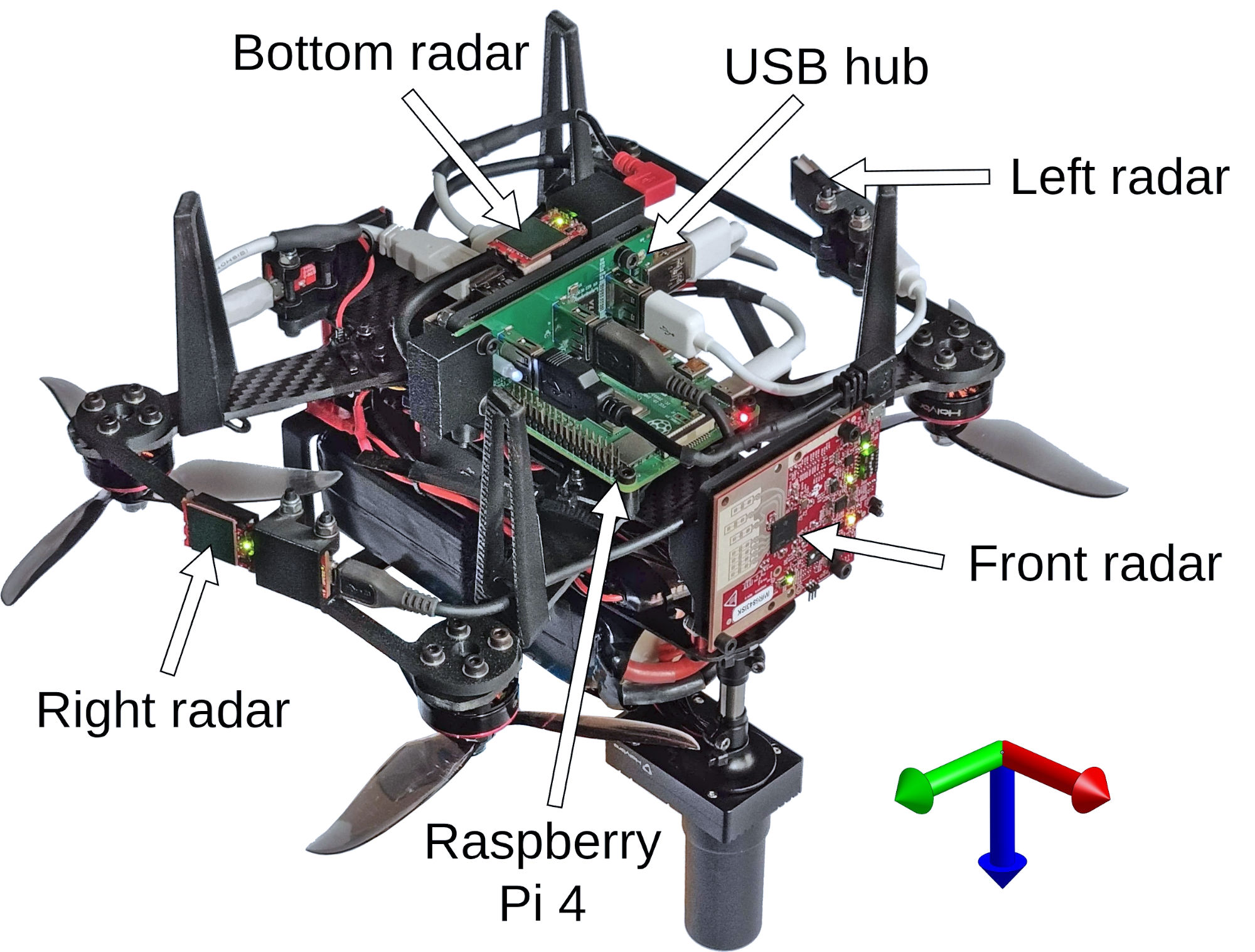}
        \caption{View of the system from the bottom front, highlighting the locations of the USB hub, Raspberry Pi 4, and some of the radar sensors.}
        \label{subfig:bot}
    \end{subfigure}

    \caption{Layout of the physical system from two perspectives.}
    \label{fig:hw_both}
\end{figure}

The base UAV platform of the overall system (seen in Fig. \ref{fig:hw_both}) is a quadcopter Holybro QAV250 with a wheelbase of 250 mm. At the center of the frame sits a Pixhawk 4 Mini autopilot that runs the PX4 flight control software stack v.1.13.0 built from source with Real-Time Publish Subscribe (RTPS) functionality. A GNSS receiver with Real-Time Kinematic Positioning (RTK) capabilities sits on top of the frame. The RTK functionality provides highly accurate ($\pm$2 cm) positioning and is used for obtaining ground truth position data during testing. However, the system has been successfully tested without RTK. 

Surrounding the UAV frame are six Texas Instruments mmWave radar devices pointing mostly along the $\pm X$, $\pm Y$ and $\pm Z$ directions. The front $+X$ radar device is a IWR6843ISK 60-64 GHz long-range antenna mmWave sensor\cite{iwr6843isk}. A long range device was chosen for the front facing direction because this is the flight direction along which the highest speeds are typically travelled. The antenna provides 120° azimuth field of view (FoV) and 30° elevation FoV, and the device is oriented such that the wide azimuth FoV sweeps up and down ($\pm Z$) while the narrower elevation FoV sweeps horizontally ($\pm Y$). While flying forward at high speeds, the UAV assumes a steep pitch angle. This specific sensor orientation makes sure that targets horizontally in front of the UAV remain within the front sensor FoV during high speed flight.

\begin{figure*}[t]
    \centering
    \includegraphics[width=0.99\linewidth]{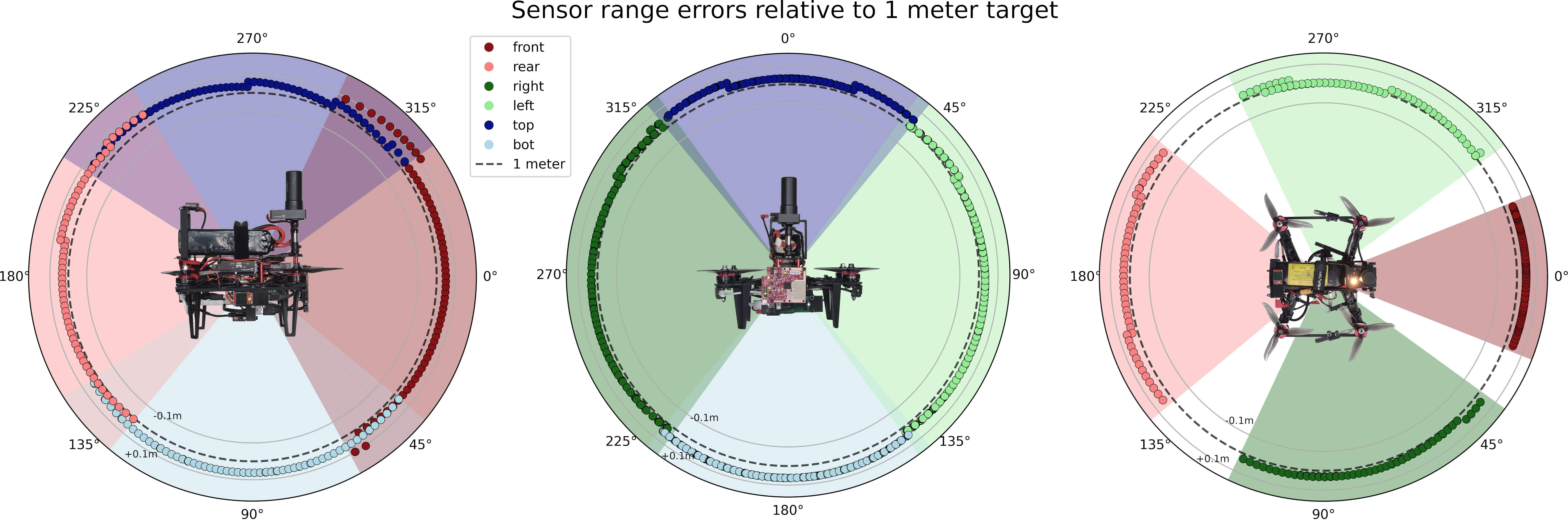}
    \caption{Visualization of the radar sensors' effective FoV and measurement errors relative to a 1 meter target in the $XZ$ (left), $YZ$ (middle), and $XY$ (right) planes. Dots represent measurements and shaded regions show the resulting effective FoV. Shades of regions with overlapping FoVs are a blend of the two overlapping shades.}
    \label{fig:meter_error}
\end{figure*}

The five other radar devices are IWR6843AOPEVM 60-64 GHz antenna-on-package (AoP) mmWave sensors\cite{iwr6843aopevm}. These are shorter range devices with a greater FoV - 120° azimuth FoV and 120° elevation FoV. To save space and weight, these devices have been reduced to just the mission AOP board by removing the optional breakaway board. The rear ($-X$), top ($+Z$), and bottom ($-Z$) sensors have their boresight aligned with the frame directions. To make up for the front sensor's narrower horizontal FoV, the left and right sensors are both angled 15° towards the front. Relative to the front sensor at 0° in the XY plane, the left sensor boresight angle is therefore 75°, and the right sensor -75°. See Fig. \ref{fig:meter_error} for the sensors' effective FoVs.

All radar sensors run the out-of-box firmware and operate in 3D mode (3RX, 4TX) with a data rate of 10 Hz. A custom ROS2 node\cite{iwr6843ros2} interfaces with the devices.

The top, left, right, and front sensors connect via USB to an externally powered USB hub for both power and data. This hub then connects via USB to a Raspberry Pi 4 which runs Ubuntu 20.04 with ROS2 Foxy as the middleware. The rear and bottom sensors are externally powered and both connect directly via USB to the Raspberry Pi for data. A USB-to-serial device connects the Pixhawk Mini autopilot to the Raspberry Pi. Power from the 4S LiPo battery is regulated to 5 V with two DCDC converters; one powering all six sensors, and the other powering the Raspberry Pi. 

\subsection{Validation of omnidirectional sensing}

With the hardware completed, multiple tests were conducted to evaluate the effective FoVs of the sensors as well as the accuracy of the distance measurements. Three separate test configurations were set up, one for testing inside each plane ($XZ$, $YZ$, $XY$, i.e. left, middle, right in Fig. \ref{fig:meter_error}). For a given test, the UAV was placed on a turntable such that the normal of the plane of interest was parallel with the gravity vector. For example, the UAV should be placed on the turntable such that its $Y$-axis is parallel to gravity in order to test in the $XZ$-plane, and the axis of rotation of the turntable should be exactly aligned with the UAV's $Z$-axis. A metallic corner reflector target is positioned on a thin wooden stick exactly 1 meter from the axis of rotation. As the UAV is slowly rotated, measurements from all six radar sensors is collected and transformed into the UAV frame. For any of the three tested planes, only four sensors detect the target during a full revolution. Fig. \ref{fig:meter_error} plots the collected data in the three planes.  From the plot it is evident that all sensors are reasonably accurate to within a few centimeters. Tab. \ref{tab:combined-sensor-stats} contains a breakdown of the results.

\begin{table}[htbp]
\centering
\small
\caption{Combined summary of the distance measurement errors of the three tests. All values in meters.}
\label{tab:combined-sensor-stats}
\sisetup{
  detect-weight = true,
  detect-inline-weight = math,
  round-mode = places,
  round-precision = 4,
  table-number-alignment = center
}
\begin{tabular}{
  l
  S[table-format=1.4]   
  S[table-format=1.4]   
  S[table-format=1.4]   
  S[table-format=1.4]   
  S[table-format=1.4]   
}
\toprule
Sensor & {$\mu$} & {$\sigma$} & {Min} & {Max} & {RMSE} \\
\midrule
Top    & 0.0603 & 0.0173 &  0.0262 & 0.0907 & 0.0628 \\
Bottom & 0.0810 & 0.0116 &  0.0510 & 0.0940 & 0.0819 \\
Left   & 0.0326 & 0.0186 & -0.0004 & 0.0593 & 0.0375 \\
Right  & 0.0426 & 0.0106 &  0.0145 & 0.0704 & 0.0439 \\
Rear   & 0.0627 & 0.0233 &  0.0079 & 0.1116 & 0.0669 \\
Front  & 0.0785 & 0.0318 &  0.0340 & 0.1136 & 0.0847 \\
\specialrule{0.08em}{0.5em}{0.5em} 
\textbf{Overall} & \textbf{0.0607} & \textbf{0.0279} & \textbf{-0.0004} & \textbf{0.1136} & \textbf{0.0663} \\
\bottomrule
\end{tabular}
\end{table}

An average distance measurement error of around 6 cm is acceptable in an obstacle avoidance context. Further testing is required to evaluate if this is a constant offset, a percentage offset, or differently scaling error.

With the same data it is also possible to estimate the actual fields-of-view of the six sensors. Tab. \ref{tab:sensor-fov} shows the estimated FoVs of the radar sensors.


\begin{table}[htbp]
\centering
\small
\caption{Measured vs.\ expected sensor fields-of-view (FoV). All values in degrees (°).}
\label{tab:sensor-fov}
\sisetup{
  detect-weight = true,
  detect-inline-weight = math,
  round-mode = places,
  round-precision = 0,
  table-number-alignment = center
}
\begin{tabular}{
  l
  @{\hskip 8pt} S[table-format=3.0] @{\hskip 4pt} S[table-format=3.0]
  @{\hskip 16pt} S[table-format=3.0] @{\hskip 4pt} S[table-format=3.0]
}
\toprule
Sensor
  & \multicolumn{2}{c}{Elevation} & \multicolumn{2}{c}{Azimuth} \\
\cmidrule(lr){2-3} \cmidrule(lr){4-5}
  & {Measured} & {Expected} & {Measured} & {Expected} \\
\midrule
Top    & 108 & 120 &  76 & 120 \\
Bottom & 106 & 120 &  76 & 120 \\
Left   & 104 & 120 &  75 & 120 \\
Right  & 104 & 120 &  75 & 120 \\
Rear   & 135 & 120 &  75 & 120 \\
Front  &  41 &  30 & 122 & 120 \\
\bottomrule
\end{tabular}
\end{table}

Curiously, all of the five IWR6843AOPEVM devices exhibit significantly narrower-than-expected azimuth FoVs and slightly narrower elevation angles. Only the rear sensor's measured elevation FoV is wider than the expected value. On the other hand, the front sensor shows slightly better measured elevation and azimuth angles compared to the expected FoVs. The result of these unexpected measured FoVs is that narrow blind spots show up when looking at the results in the $XY$-plane. A simple way to address this would be to substitue the IWR6843ISK device with another IWR6843AOPEVM, sacrificing long-range detection for a wider FoV. However, there is also the chance that the observed deviation is somehow due to the corner reflector target used during these tests. As shown later, testing with real powerlines seems to greatly diminish this discrepancy.

\section{Radar Behavior in Power Line Environments}
\label{sec:radar_pl_behavior}
\begin{figure}[b]
	\centering
	\includegraphics[width=0.85\linewidth]{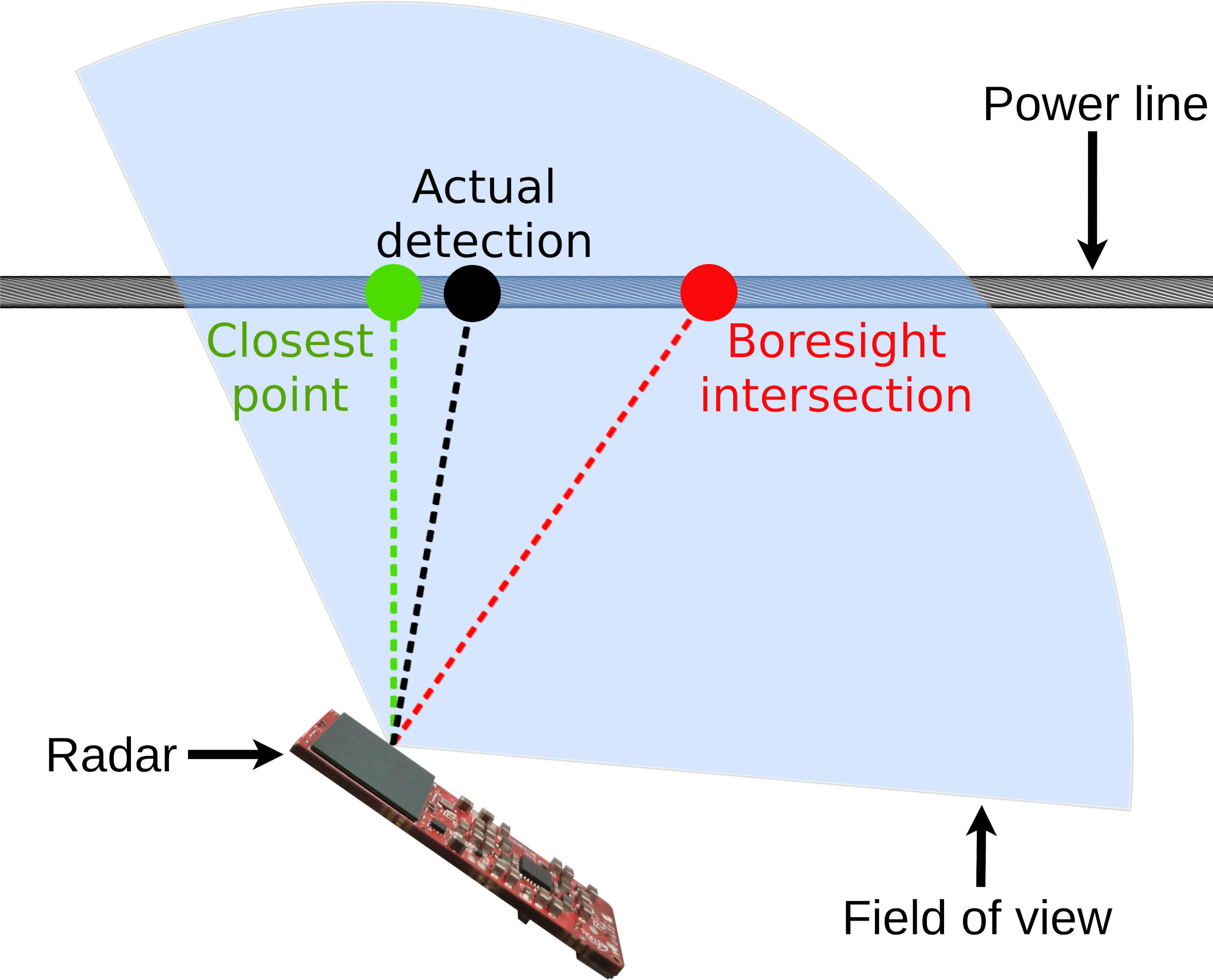}
 	\caption{When pointed at an angle towards a power line, where does the radar sensor detect the power line?}
	\label{fig:pl_angle}
\end{figure}

To build a reliable power line detection and avoidance algorithm requires knowledge on the behavior of mmWave radars in such environments. In particular one question needs answering: which point on the power line does the radar detect? Fig. \ref{fig:pl_angle} illustrates the situation. A mmWave radar device points toward a power line. However, the boresight of the radar is not perpendicular to the power line, i.e. there is some non-zero angle between the boresight and the vector from radar to the closest point on the power line. The question now is: \textit{which point does the radar detect?}

To find out, a simple test is performed. While hovering in front of a power line the UAV is commanded to slowly yaw while recording data from all sensors as well as the relative rotation of the UAV to the power line. For each detection of the power line in the collected data, the angle between the detected point (black in Fig. \ref{fig:pl_angle}) and closest point (green in Fig. \ref{fig:pl_angle}) can now be compared to the angle between the boresight of the radar from which the detection originated (red in Fig. \ref{fig:pl_angle}) and the closest point. 

\begin{figure}[h]
  \centering
  \includegraphics[width=0.99\linewidth]{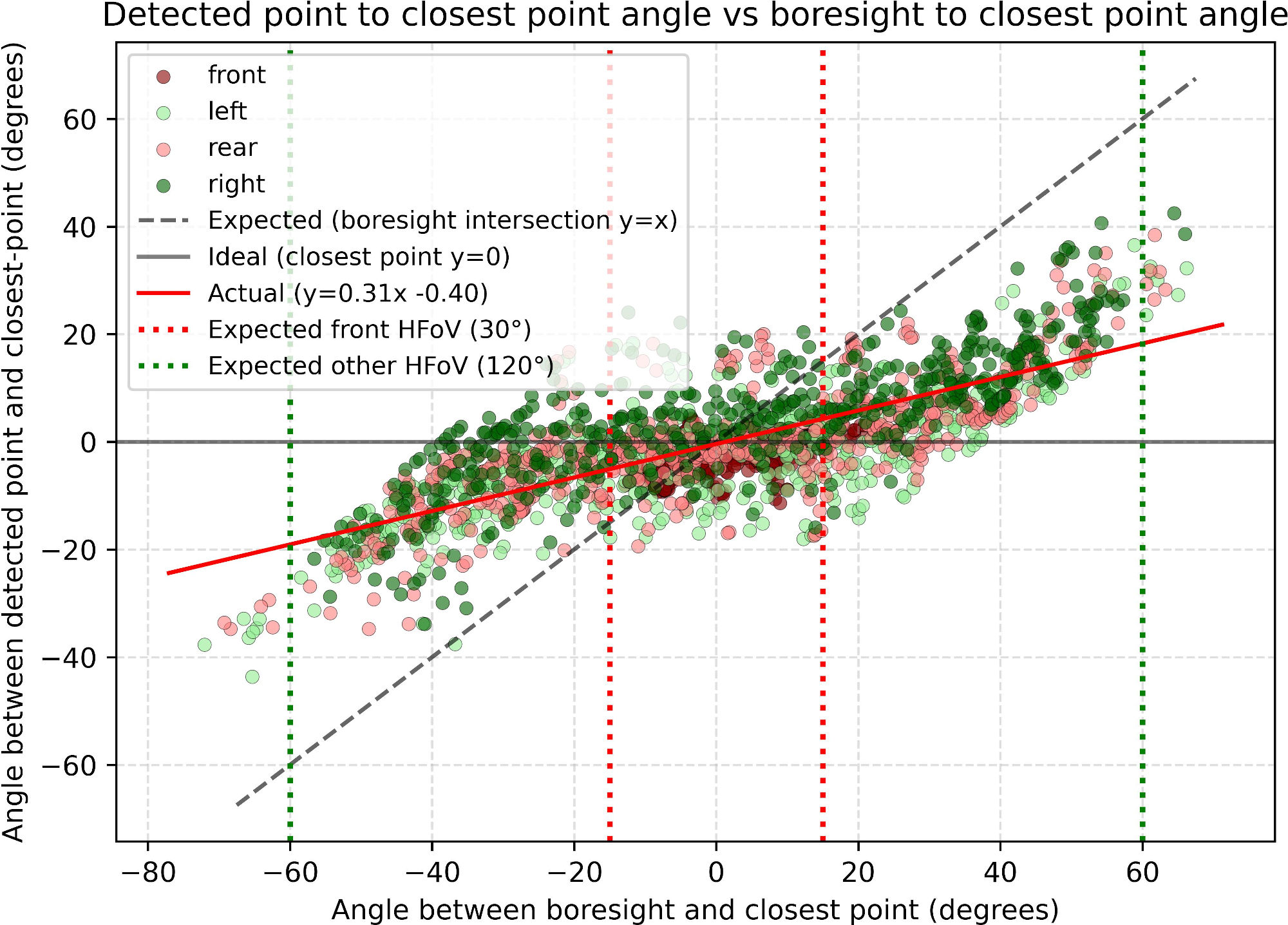}
  \caption{Plot showing the relationship of the angle between detected and closest point vs angle between boresight and closest point. While the radar boresight is still relatively close ($\pm$30°) to being perpendicular to the power line, the detected point is approximately equal to the closest point on the power line. As the boresight becomes even less perpendicular to the power line, the detected point moves further towards the boresight intersection with the power line.}
  \label{fig:angular_error}
\end{figure}

Fig. \ref{fig:angular_error} shows the relationship between the two angles. The plot highlights an interesting phenomenon (from here on referred to as $P_a$): as long as the radar boresight is still relatively close ($\pm$30°) to being perpendicular to the power line, the detected point is approximately equal to the closest point on the power line. Getting the closest point on the power line with just a single measurement and without needing to do any processing is valuable and can be used to greatly simplify collision avoidance processing. Even though the effect of $P_a$ begins to subside as the boresight becomes less perpendicular to the power line, the rest of this work assumes that detections of power lines are approximately equal to the closest point on the power line.

\section{Power Line Avoidance}
\label{sec:avoidance}
\begin{figure}[h]
	\centering
	\includegraphics[width=0.99\linewidth]{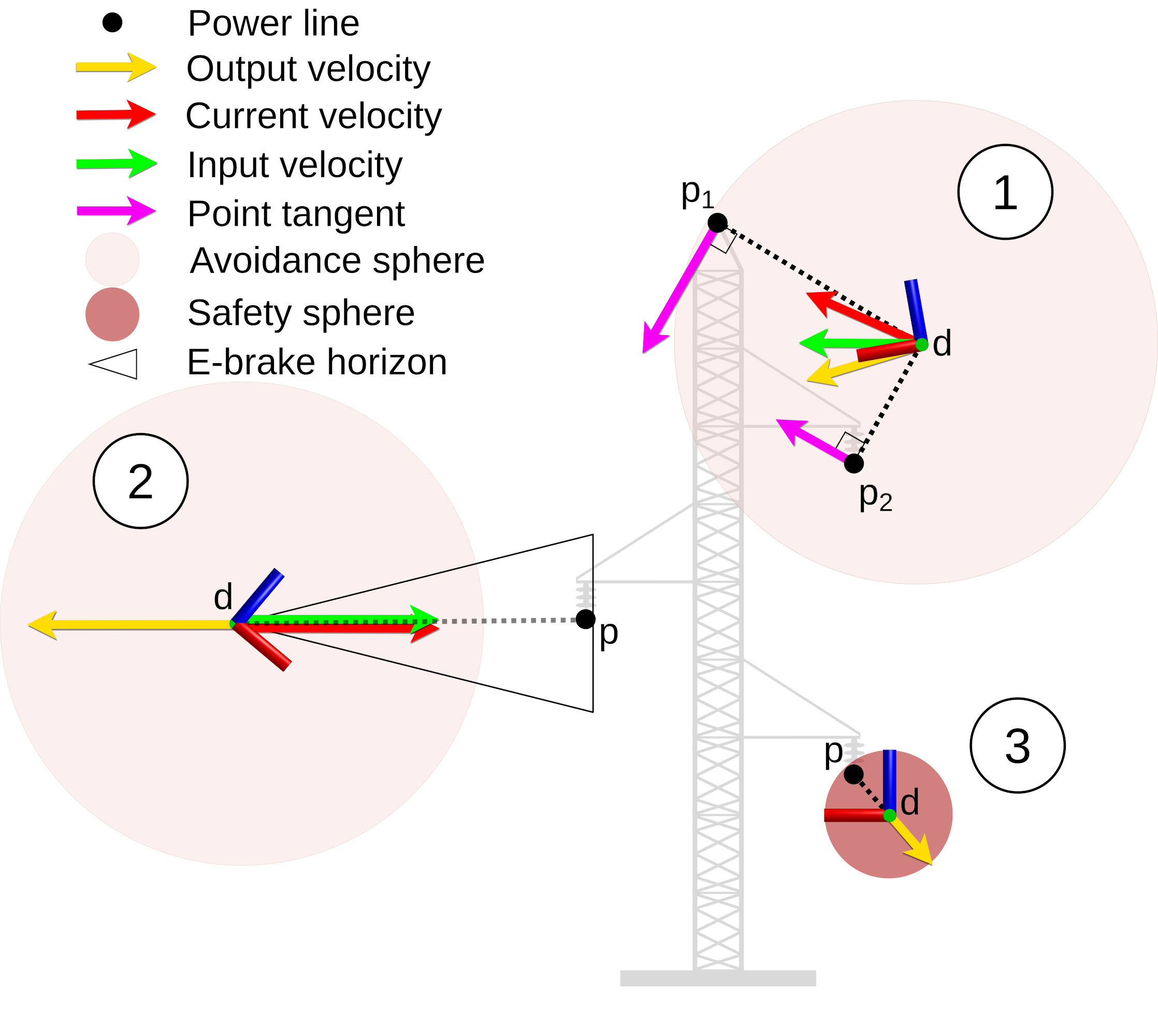}
 	\caption{Three scenarios of power line avoidance in the plane $K_p$. \textbf{1)} When the UAV travels at medium speeds, tangents to detections are used to steer UAV. \textbf{2)} At high speeds, detections within the e-brake horizon immediately slows the drone to medium speeds. \textbf{3)} Very close detections triggers a retreat away form the detection.}
	\label{fig:calculations}
\end{figure}

Given $P_a$, the avoidance problem can be reduced to two dimensions. Since power lines are roughly straight (at least within the detection radius of the UAV), and all conductors are typically parallel, any detection $\boldsymbol{p}_i$ will lie approximately on the plane $K_p$ that is perpendicular to the unit direction of the power lines $\boldsymbol{\hat{c}}$ and contains the UAV position $\boldsymbol{d}$. For any three dimensional vector, e.g. the UAV velocity $\boldsymbol{v}_d$, obtained from the autopilot's state vector $\boldsymbol{x}(t)$, only their components within $K_p$ can be considered. Fig. \ref{fig:calculations} shows the simplified scenario, i.e. the projection onto $K_p$.

\subsection{Scenario 1: Tangential avoidance}

While the UAV is not in immediate risk of collision, power line avoidance is based on calculating tangents to the detections and using those to steer the UAV around the power lines. Corrections are calculated once with both the UAV's current velocity $\boldsymbol{v}_d$ as well as with the user desired velocity $\boldsymbol{v}_u$ (e.g. from a pilot or autonomy stack). For brevity, only calculations with $\boldsymbol{v}_d$ are shown here. 

The process begins by filtering points based on the avoidance sphere. This is a sphere around the UAV with radius $r_a$ inside which any detections are considered for avoidance.


\begin{equation}
\begin{gathered}
        ||\boldsymbol{p}_i|| < r_a
\end{gathered}
\label{eq:1}
\end{equation}

The next step is to determine if a correction is needed, i.e. if $\boldsymbol{v}_d$ has a component towards detection $\boldsymbol{p}_i$:

\begin{equation}
\begin{gathered}
        \boldsymbol{\hat{p}}_i \cdot \boldsymbol{\hat{v}}_d > 0
\end{gathered}
\label{eq:2}
\end{equation}

where $\boldsymbol{\hat{p}}_i$ and $\boldsymbol{\hat{v}}_d$ are unit vectors. The detection tangent vector $\boldsymbol{t}_i$, which is orthogonal to $\boldsymbol{\hat{p}}_i$, can then be found by first crossing the detection unit vector $\boldsymbol{\hat{p}}_i$ with the negative gravity unit vector $\boldsymbol{\hat{g}}_n$ and then crossing the result of that with $\boldsymbol{\hat{p}}_i$:

\begin{equation}
\begin{gathered}
        \boldsymbol{t}_i = ( \boldsymbol{\hat{p}}_i \times \boldsymbol{\hat{g}}_n )\times \boldsymbol{\hat{p}}_i
\end{gathered}
\label{eq:3}
\end{equation}

The dot product between the tangent $\boldsymbol{t}_p$ and UAV velocity $\boldsymbol{v}_d$ can then be used to determine which tangent direction $\boldsymbol{\hat{t}}_{i,f}$ to go around the detected power line $\boldsymbol{p}_i$

\begin{equation}
\begin{gathered}
    \mathbf{\hat{t}}_{i,f} =
    \begin{cases}
    \hat{\mathbf{t}}_i, & \text{if } \mathbf{t}_i\cdot\boldsymbol{v}_d > 0,\\[4pt]
    -\hat{\mathbf{t}}_i, & \text{otherwise,}
    \end{cases}
\end{gathered}
\label{eq:4}
\end{equation}

Finally, the magnitude of the tangent is scaled by the degree of parallelity between the unit tangent $\boldsymbol{\hat{t}}_{i,f}$ and the unit drone velocity $\boldsymbol{\hat{v}}_d$.

\begin{equation}
\begin{gathered}
    \boldsymbol{t}_{i,s} = ( \boldsymbol{\hat{p}}_i \cdot \boldsymbol{\hat{v}}_d ) \boldsymbol{\hat{t}}_{i,f} 
\end{gathered}
\label{eq:5}
\end{equation}

The closer the unit drone velocity $\boldsymbol{\hat{v}}_d$ is to pointing directly towards the detected power line $\boldsymbol{\hat{p}}_i$, the larger the magnitude of the final tangent $\boldsymbol{t}_{i,s}$ will be.

Eqs. \ref{eq:1}-\ref{eq:5} are then performed once more, this time with user desired velocity $\boldsymbol{v}_u$ substituting the drone velocity $\boldsymbol{v}_d$ to obtain a second tangent for that detection. Eqs. \ref{eq:1}-\ref{eq:5} are then repeated for every detection with both $\boldsymbol{v}_d$ and $\boldsymbol{v}_u$. The resulting set of tangents is then summed into a single tangent $\boldsymbol{t}_{s}$. $\boldsymbol{t}_{s}$ is then summed with $\boldsymbol{v}_u$, and this vector's magnitude is clamped to that of $\boldsymbol{v}_u$ to create the final corrected output velocity $\boldsymbol{v}_{out}$.


\begin{equation}
\boldsymbol{v}_{\mathrm{out}} =
\begin{cases}
\displaystyle \min\!\left(1,\ \frac{\|\boldsymbol{v}_u\|}{\|\boldsymbol{t}_s+\boldsymbol{v}_u\|}\right)\,(\boldsymbol{t}_s+\boldsymbol{v}_u), 
& \text{if } \|\boldsymbol{t}_s+\boldsymbol{v}_u\|>0,\\[10pt]
\boldsymbol{0}, & \text{if } \|\boldsymbol{t}_s+\boldsymbol{v}_u\|=0.
\end{cases}
\end{equation}

\begin{figure}[b]
	\centering
	\includegraphics[width=0.99\linewidth]{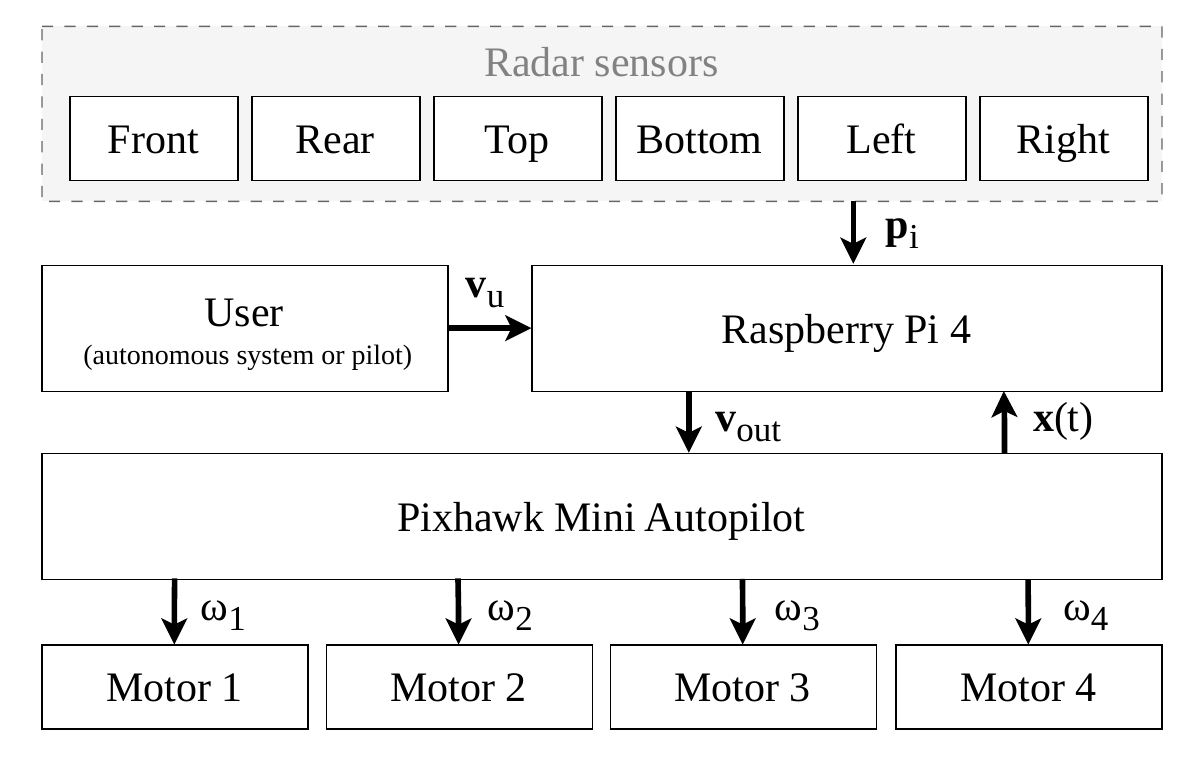}
 	\caption{Data flow in the power line avoidance system.}
	\label{fig:dataflow}
\end{figure}

This value is then sent to the autopilot. The overall data flow is shown in Fig. \ref{fig:dataflow}.

\subsection{Scenario 2: High speed}

The second scenario, as also illustrated in Fig. \ref{fig:calculations}, is related to avoiding power lines while flying at greater speeds. Under such circumstances, an e-brake horizon cone $H$ expands outward in the direction of $\boldsymbol{v}_d$. As $||\boldsymbol{v}_d||$ increases, so does the length of $H$. The length $l_H$ of the cone is defined as

\begin{equation}
\begin{gathered}
    l_H = \frac{||\boldsymbol{v}_d||^2}{2a_{max}}+s_{margin}
\end{gathered}
\label{eq:7}
\end{equation}

where $a_{max}$ is the maximum acceleration of the UAV and $s_{margin}$ is a safety margin. At the far end, the cone is defined to have a length-to-width ratio of 2:1, yielding an angle $\alpha$ of $\text{atan}(1/4)=0.245$ radians from center to diagonal. When the UAV moves above a certain speed $\boldsymbol{v}_{eb}$, each detected point $\boldsymbol{p}_i$ is checked if it is within the e-brake horizon.

\begin{equation}
\begin{gathered}
    l_H > ||\boldsymbol{p}_i||
\end{gathered}
\label{eq:9}
\end{equation}

\begin{equation}
\begin{gathered}
    \alpha > \text{atan2}(||\boldsymbol{v}_d\times\boldsymbol{p}_i||,\boldsymbol{v}_d\cdot\boldsymbol{p}_i)
\end{gathered}
\label{eq:10}
\end{equation}

If Eqs. \ref{eq:9} and \ref{eq:10} are both true, immediate maximum braking is commanded. Braking continues until the UAV no longer moves faster than $\boldsymbol{v}_{eb}$ at which point the situation turns into scenario 1 or 3.

\subsection{Scenario 3: Close proximity}

\begin{figure*}[b]
  \centering
  \begin{tabular}{@{}p{\colA}@{\hspace{\hgap}}p{\colB}@{\hspace{\hgap}}p{\colC}@{}}
    \multicolumn{2}{@{}p{\dimexpr\colA+\colB+\hgap\relax}@{}}{%
      \centering
      \begin{subfigure}[b]{\linewidth}
        \centering
        \includegraphics[width=\linewidth,height=\rowA,keepaspectratio]{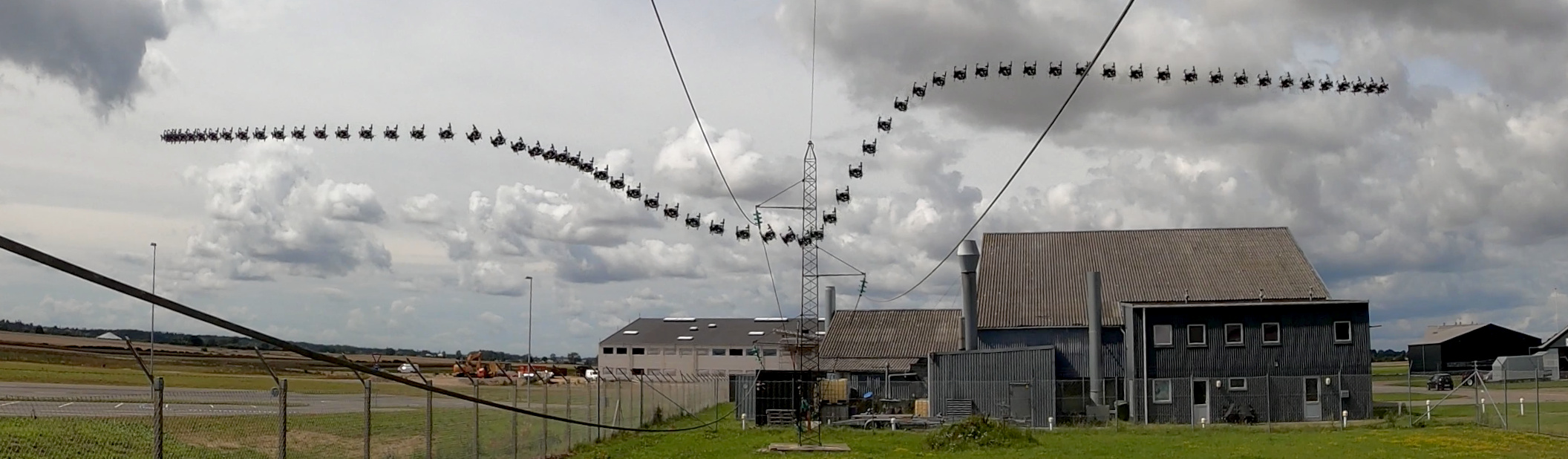}
        \caption{Power lines are avoided (left to right) while flying at upwards of 10 m/s. The top wire is 10 mm in diameter, while the bottom three conductors are 20 mm in diameter. The shortest distance between the power lines is about 3 meters.}
        \label{fig:top-wide}
      \end{subfigure}
    } &
    \multirow{2}{*}{%
      \begin{minipage}[c][\dimexpr\rowA+\vgap+\rowB\relax][c]{\colC}
        \centering
        \begin{subfigure}[b]{\linewidth}
          \centering
          \includegraphics[width=\linewidth,height=\rightHeight]{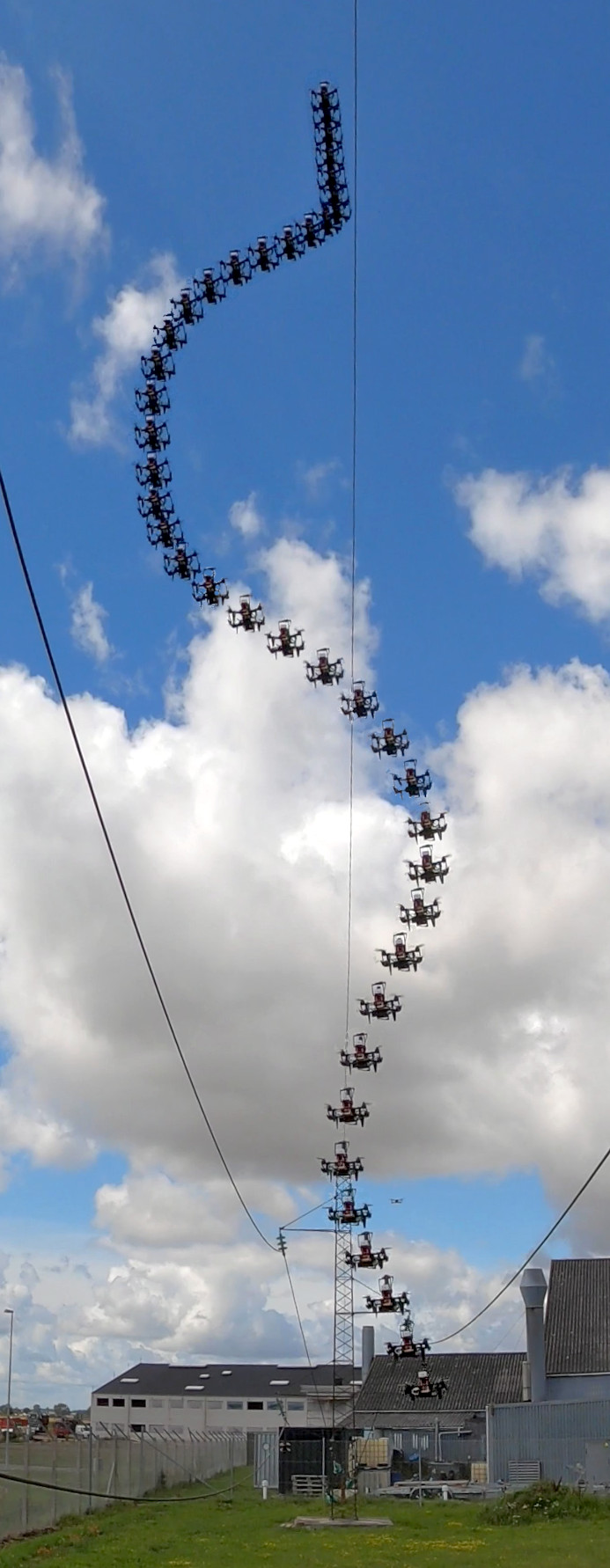}
          \caption{The UAV avoids multiple power lines while descending.}
          \label{fig:right-tall}
        \end{subfigure}
        \vspace{106mm}
      \end{minipage}
    } \\
    \parbox[b][\rowB][c]{\colA}{\centering
      \begin{subfigure}[b]{\linewidth}
        \centering
        \includegraphics[width=\linewidth,height=\rowB,keepaspectratio]{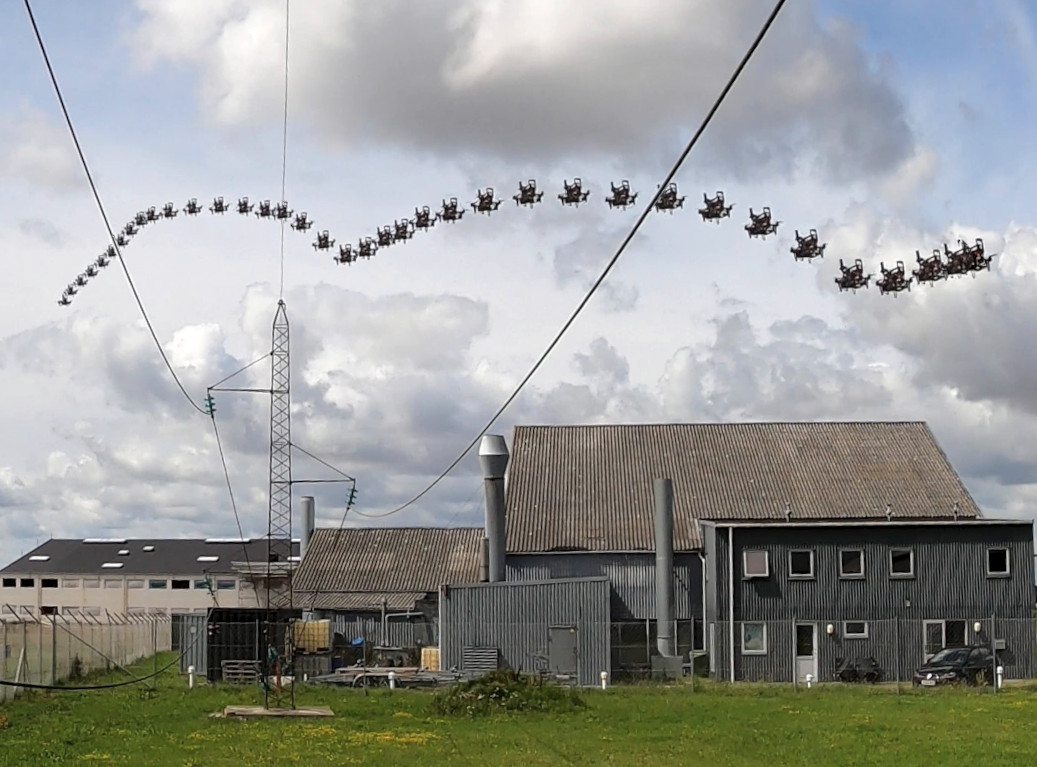}
        \caption{Power lines are avoided even when approached from an oblique angle (moving right to left).}
        \label{fig:bottom-left}
      \end{subfigure}
    } &
    \parbox[b][\rowB][c]{\colB}{\centering
      \begin{subfigure}[b]{\linewidth}
        \centering
        \includegraphics[width=\linewidth,height=\rowB,keepaspectratio]{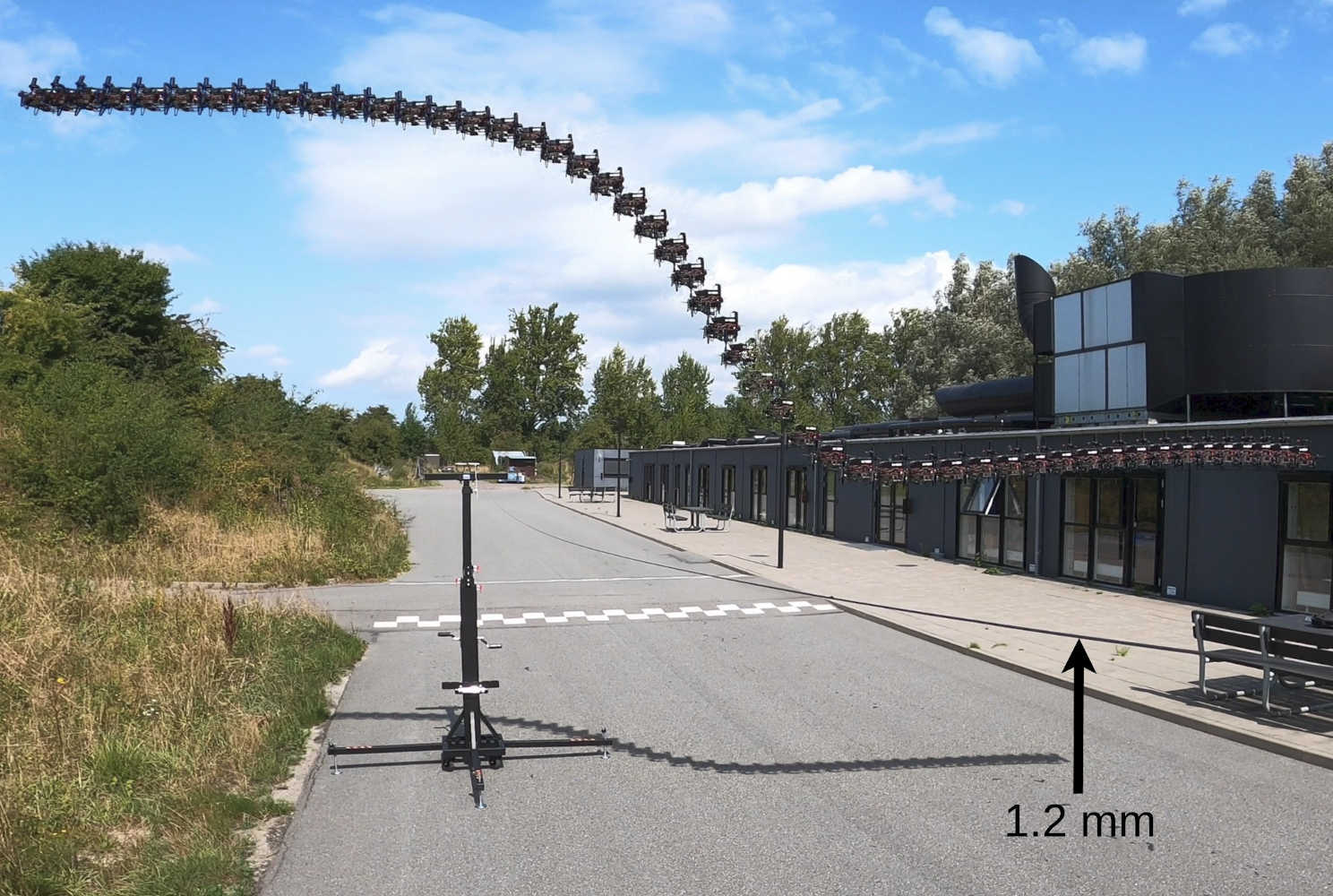}
        \caption{The system is able to detect and avoid a 1.2 mm diameter steel wire (moving right to left).}
        \label{fig:bottom-middle}
      \end{subfigure}
    } &
    \\[\vgap]
  \end{tabular}

  \caption{Four examples of the system performing power line avoidance. Three of the examples are performed in a decommissioned stretch of power line, while the fourth example (d) showcases the system's ability to avoid even very thin wires of just 1.2 mm in diameter. These tests and more are also shown in the supplemental video demonstration\cite{video_demo}.}
  \label{fig:2x3block}
\end{figure*}

As indicated in Fig. \ref{fig:calculations}, scenario 3 is straight forward; if a power line is detected within a safety sphere, the UAV immediately rejects itself from that detection. The safety sphere is similar to the avoidance sphere but significantly smaller and defined by its radius $r_s$. All detections are checked against the safety sphere.

\begin{equation}
\begin{gathered}
        ||\boldsymbol{p}_i|| < r_s
\end{gathered}
\label{eq:11}
\end{equation}

If Eq. \ref{eq:11} equates to true, a corrective velocity $\boldsymbol{v}_{out}$ is calculated as

\begin{equation}
\begin{gathered}
        \boldsymbol{v}_{out} = k_s(||\boldsymbol{p}_i||-r_s)\boldsymbol{\hat{p}}_i
\end{gathered}
\label{eq:12}
\end{equation}

where $k_s$ is used to scale the intensity of the rejection. Furthermore, the closer the detection is to the center of the safety sphere, the more strongly the resulting rejection will be.

\section{Results}
\label{sec:experiments}

The functionality of the system is validated in an outdoor power line environment. The specific environment is an approximately 35 meter long stretch of decommissioned 3-phase power transmission lines which consists of three 20 mm conductors arranged in a triangle relative to each other with a single 10 mm conductor on top of the others - the same arrangement as shown in Fig. \ref{fig:calculations}. The closest distance between power lines is approximately 3 meters, and the stack is approximately 9 meters tall. Figs. \ref{fig:top-wide}, \ref{fig:right-tall} and \ref{fig:bottom-left} show the system performing power line avoidance in the described environment from various directions and angles. When approaching the power lines with the long-range front sensor, the power lines are typically detected from beyond 10 meters away, which enables the system to perform avoidance maneuvers at velocities upwards of 10 m/s. The shorter range sensors detect the power lines from approximately 7 meters, and avoidance maneuvers in the direction of these sensors can be performed at up to 5 m/s.

The tests in Figs. \ref{fig:top-wide}, \ref{fig:bottom-left} and \ref{fig:bottom-middle} were all performed by inputting a desired velocity with only horizontal components towards the power line and then continuously inputting the same desired velocity throughout the test. As the system nears a power line, the avoidance algorithm corrects the desired velocity to create an output velocity that steers around the power lines. 
In Fig. \ref{fig:top-wide}, the UAV approaches the power lines at 5 m/s from the left. As the power line is detected within the e-brake horizon, the UAV quickly brakes until its velocity is sufficiently low, at which point it continues at a constant medium speed while avoiding two power lines, going below the first and above the second.
Fig. \ref{fig:bottom-left} is a similar test but performed at an angle to the power line. The purpose is to evaluate if the blind spots seen in Fig. \ref{fig:meter_error} impact the system during real world testing. As visualized by the image sequence, the UAV successfully avoided the power lines without issue, indicating that the observed blind spots may be an artifact of the laboratory test setup.
The image sequence in Fig. \ref{fig:bottom-middle} shows the system avoiding a thin steel wire of just 1.2 mm in diameter. This was the smallest gauge wire available during testing, so the system may be able to avoid even thinner wires. This opens up the work to be applicable not only in power line environments, but also in environments where thinner wires are used, such as telephone wires.
Since the UAV is not only equipped with sensors in the horizontal plane but also pointing up and downwards, it is able to safely navigate between power lines in these directions as well. Fig. \ref{fig:right-tall} shows the UAV smoothly avoiding the wires while the only user-input is a velocity pointing straight down. This may be especially useful in situations where the UAV autonomously takes off or lands, for example during emergency landings in the vicinity of power lines.

\begin{figure}[t]
  \centering
  \includegraphics[width=0.99\linewidth]{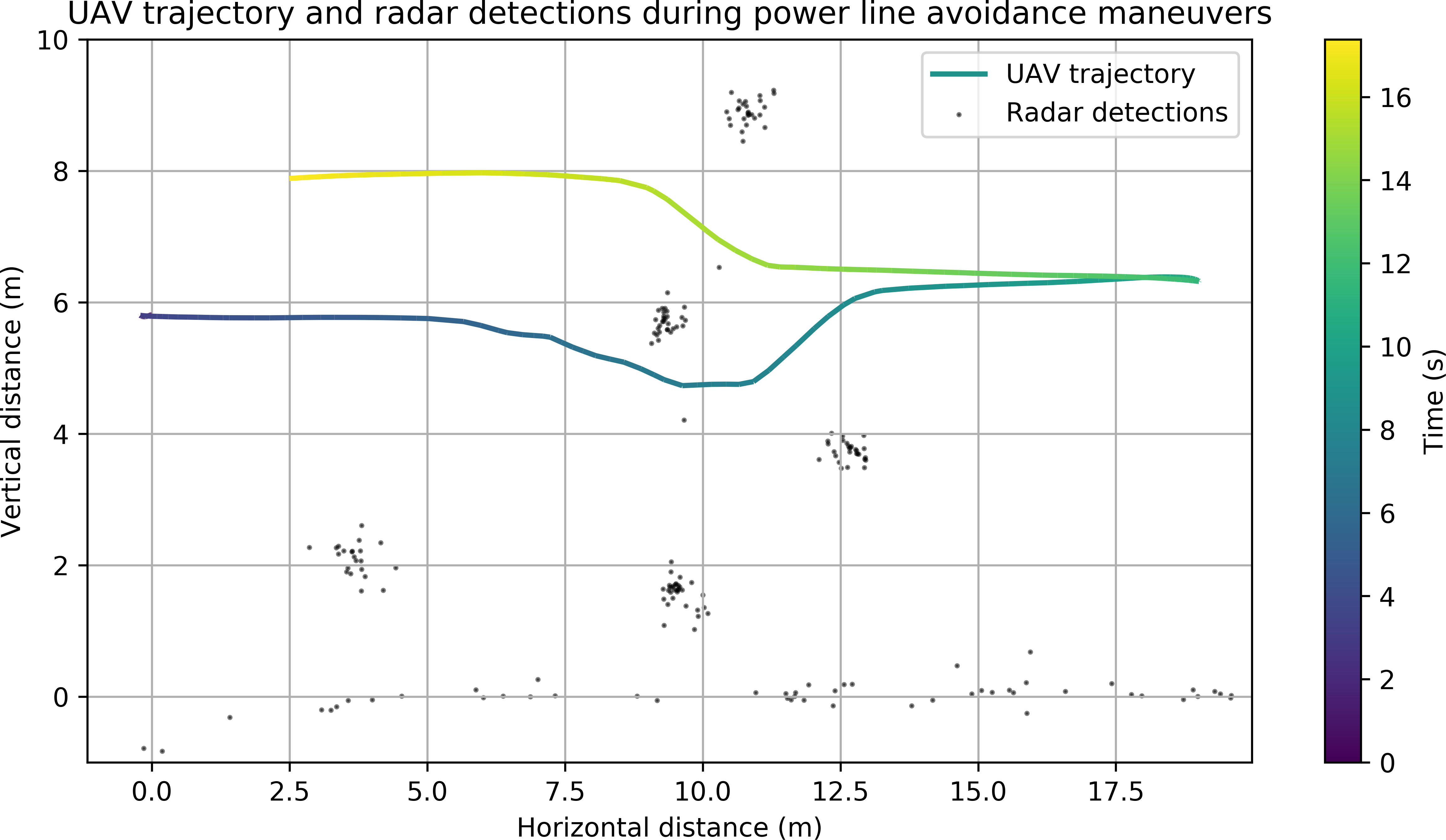}
  \caption{Plot of the UAV's trajectory around radar detections during power line avoidance maneuvers. The input velocities are entirely in the horizontal plane, and the avoidance algorithm automatically corrects the output velocity to steer the UAV around the power lines.}
  \label{fig:avoidance_plot}
\end{figure}

The plot in Fig. \ref{fig:avoidance_plot} shows the radar detections and UAV trajectory in the plane $K_p$ while avoiding power lines. The detections from various sensors are consistent throughout the full 16 second test, and the system never gets closer than 1 meter to a detected power line.


\section{Conclusion}
\label{sec:conclusion}
This work presented a purely radar-based approach to power line detection and avoidance for UAV. The multi-sensor perception system ensures omnidirectional sensing, with device specifications optimized based on typical flight patterns. The sensor suite was tested in a laboratory setting to verify its capabilities as well as with outdoors power lines to characterize its behavior in a representative setting. An interesting phenomenon was observed in which the radar sensors detect power lines at approximately the closest point to the sensor. This phenomenon was used to develop an efficient power line avoidance algorithm, which was thoroughly evaluated in a power line environment. The system proved its effectiveness by robustly detecting and avoiding power lines and even thin wires of just 1.2 mm in diameter.
\balance


\addtolength{\textheight}{-0.0cm}   
\balance

\bibliographystyle{./bibliography/IEEEtran}
\bibliography{./bibliography/refs.bib}

@ARTICLE{Malle2022,
  author={Malle, Nicolaj Haarhøj and Nyboe, Frederik Falk and Ebeid, Emad Samuel Malki},
  journal={{IEEE Access}}, 
  title={{Onboard Powerline Perception System for UAVs Using mmWave Radar and FPGA-Accelerated Vision}}, 
  year={2022},
  volume={10},
  number={},
  pages={113543-113559},
  doi={10.1109/ACCESS.2022.3217537}}

@INPROCEEDINGS{Malle2021,
  author={Malle, Nicolaj Haarhøj and Nyboe, Frederik Falk and Ebeid, Emad},
  booktitle={2021 International Conference on Unmanned Aircraft Systems (ICUAS)}, 
  title={{Survey and Evaluation of Sensors for Overhead Cable Detection using UAVs}}, 
  year={2021},
  volume={},
  number={},
  pages={361-370},
  doi={10.1109/ICUAS51884.2021.9476724}}

@INPROCEEDINGS{icra2023,
  author={Nyboe, Frederik Falk and Malle, Nicolaj Haarhøj and Bögel, Gerd vom and Cousin, Linda and Heckel, Thomas and Troidl, Konstantin and Madsen, Anders Schack and Ebeid, Emad},
  booktitle={2023 IEEE International Conference on Robotics and Automation (ICRA)}, 
  title={{Towards Autonomous UAV Railway DC Line Recharging: Design and Simulation}}, 
  year={2023},
  volume={},
  number={},
  pages={3310-3316},
  doi={10.1109/ICRA48891.2023.10161506}}

@INPROCEEDINGS{icra2024,
  author={Hoang, Viet Duong and Falk Nyboe, Frederik and Malle, Nicolaj Haarhøj and Ebeid, Emad},
  booktitle={2024 IEEE International Conference on Robotics and Automation (ICRA)}, 
  title={Autonomous Overhead Powerline Recharging for Uninterrupted Drone Operations}, 
  year={2024}}

@INPROCEEDINGS{iros2024,
  author={Malle, Nicolaj Haarhøj and Ebeid, Emad},
  booktitle={2024 IEEE/RSJ International Conference on Intelligent Robots and Systems (IROS)}, 
  title={Autonomous Power Line Tracking with mmWave Radar}, 
  year={2024}}

@INPROCEEDINGS{wessendorp2021,
  author={Wessendorp, Nikhil and Dinaux, Raoul and Dupeyroux, Julien and de Croon, Guido C. H. E.},
  booktitle={2021 IEEE/RSJ International Conference on Intelligent Robots and Systems (IROS)}, 
  title={Obstacle Avoidance onboard MAVs using a FMCW Radar}, 
  year={2021}}

@INPROCEEDINGS{9973561,
  author={Sun, Yue and Huang, Zhuoming and Zhang, Honggang and Liang, Xiaohui},
  booktitle={2022 IEEE 19th International Conference on Mobile Ad Hoc and Smart Systems (MASS)}, 
  title={3D Reconstruction of Multiple Objects by mmWave Radar on UAV}, 
  year={2022}}

@INPROCEEDINGS{9341432,
  author={Yu, Hang and Zhang, Fan and Huang, Panfeng and Wang, Chen and Yuanhao, Li},
  booktitle={2020 IEEE/RSJ International Conference on Intelligent Robots and Systems (IROS)}, 
  title={Autonomous Obstacle Avoidance for UAV based on Fusion of Radar and Monocular Camera}, 
  year={2020}}

@Article{drones6110361,
AUTHOR = {Bigazzi, Luca and Miccinesi, Lapo and Boni, Enrico and Basso, Michele and Consumi, Tommaso and Pieraccini, Massimiliano},
TITLE = {Fast Obstacle Detection System for UAS Based on Complementary Use of Radar and Stereoscopic Camera},
JOURNAL = {Drones},
YEAR = {2022}
}

@Article{radarcameracnn,
AUTHOR = {Wang, Xiyue and Wang, Xinsheng and Zhou, Zhiquan and Song, Yanhong},
TITLE = {Fast detection and obstacle avoidance on UAVs using lightweight convolutional neural network based on the fusion of radar and camera},
JOURNAL = {Applied Intelligence},
YEAR = {2024}
}

@ARTICLE{10772391,
  author={Wu, Sheng-Wei and Fu, Zi-Hao and Shen, Chang-Shao and Lin, Kun-You and Chen, Shih-Yuan},
  journal={IEICE Transactions on Communications}, 
  title={A Beam-Switching Planar Antenna Module with 360° In-Plane Coverage for Drone Collision Avoidance Radar}, 
  year={2025}}

@ARTICLE{10075053,
  author={Milias, Christos and Andersen, Rasmus B. and Muhammad, Bilal and Kristensen, Jes T. B. and Lazaridis, Pavlos I. and Zaharis, Zaharias D. and Mihovska, Albena and Hermansen, Dan D. S.},
  journal={IEEE Transactions on Intelligent Transportation Systems}, 
  title={UAS-Borne Radar for Autonomous Navigation and Surveillance Applications}, 
  year={2023}}

@online{iwr6843aopevm,
    author        =     {{Texas Instruments}},
    title         =     {{"IWR6843AOPEVM product page"}},
    url           =     "https://www.ti.com/tool/IWR6843AOPEVM",
    note          =     {Visited on 05/09/2025}
}

@online{iwr6843isk,
    author        =     {{Texas Instruments}},
    title         =     {{"IWR6843ISK product page"}},
    url           =     "https://www.ti.com/tool/IWR6843ISK",
    note          =     {Visited on 05/09/2025}
}

@online{iwr6843ros2,
    author        =     {{Github}},
    title         =     {{"IWR6843* ROS2 node"}},
    url           =     "https://github.com/nhma20/xwr6843_ros2",
    note          =     {Visited on 05/09/2025}
}

@online{video_demo,
    author        =     {{YouTube}},
    title         =     {{"Video demonstration"}},
    url           =     "https://www.youtube.com/watch?v=rJW3eEC-5Ao",
    note          =     {Visited on 05/09/2025}
}

@online{srd_ros2_pkg,
    author        =     {{Github}},
    title         =     {{"ROS2 package repository"}},
    url           =     "https://github.com/nhma20/spherical-radar-drone",
    note          =     {Visited on 05/09/2025}
}


\begin{textblock*}{\paperwidth-40mm}(20mm,260mm)
  \centering\small
  Accepted for publication at IEEE International Conference
on Robotics and Automation (ICRA) 2024.
© 2024 IEEE. Personal use of this material is permitted.
Permission from IEEE must be obtained for all other uses, in
any current or future media, including reprinting/republishing
this material for advertising or promotional purposes, creating new collective works, for resale or redistribution to
servers or lists, or reuse of any copyrighted component of
this work in other works.
\end{textblock*}

\end{document}